\documentclass[sigconf]{acmart}

\usepackage{booktabs} 
\usepackage{multirow}

\setcopyright{rightsretained}

\acmConference{2017 AdKDD \& TargetAd}{August 2017}{Halifax, Canada}

\begin{document}
\title{Cost-sensitive Learning for Utility Optimization in\\ Online Advertising Auctions}

\author{Flavian Vasile}
\affiliation{
  \institution{Criteo}
  \city{Paris}
  \state{France}}
\email{f.vasile@criteo.com}

\author{Damien Lefortier}
\authornote{Work was done while at Criteo.}
\affiliation{
  \institution{Facebook}
  \city{London}
  \state{UK}}
\email{dlefortier@fb.com}

\author{Olivier Chapelle}
\authornotemark[1]
\affiliation{
  \institution{Google}
  \city{Mountain View}
  \state{CA}}
\email{olivier@chapelle.cc}

\renewcommand{\shortauthors}{F. Vasile et al.}

\def\x{\mathbf x}
\def\w{\mathbf w}

\begin{abstract}
One of the most challenging problems in computational advertising is the prediction of click-through and conversion rates for bidding in online advertising auctions. An unaddressed problem in previous approaches is the existence of highly non-uniform misprediction costs. While for model evaluation these costs have been taken into account through recently proposed business-aware offline metrics -- such as the {\em Utility} metric which measures the impact on advertiser profit -- this is not the case when training the models themselves. In this paper, to bridge the gap, we formally analyze the relationship between optimizing the Utility metric and the log loss, which is considered as one of the state-of-the-art approaches in conversion modeling. Our analysis motivates the idea of weighting the log loss with the business value of the predicted outcome. We present and analyze a new cost weighting scheme and show that significant gains in offline and online performance can be achieved.
\end{abstract}

%
%

\ccsdesc[500]{Information systems~Online advertising}
\ccsdesc[500]{Computing methodologies~Machine learning approaches}

\keywords{Display advertising; machine learning; conversion prediction}

\maketitle

\section{Introduction}
\label{intro:costsensitivelearning}

Online advertising is becoming a large part of the global marketing reaching \$170 billion revenue in 2015 \cite{statista}. Depending on the goal of the advertising campaign, different pricing schemes exist, but out of them, brand and performance advertising are the most prevalent.
Brand advertising is used by advertisers that want to maximize the exposure of their advertising message to online users and is priced in terms of number of ad impressions, with the cost usually referred as CPM (cost-per-mille). 
By contrast, performance advertising is appealing to advertisers that are interested in reaching certain measurable goals such as increased number of visits to their websites, increased number of leads, sales or downloads. In this case, the cost is referred as CPC (cost-per-(ad)click) or CPA (cost-per-conversion).

The marketplace that makes online advertising possible is roughly formed out of three types of players, namely the advertiser (the demand of ad display opportunities), the publisher (the offer of ad display opportunities) and the auction house, represented by a Real-Time Bidding (RTB) platform. Most of the RTB platforms use a 2nd price model \cite{yuan2013real}, where advertisers or agents representing the advertisers bid for display opportunities, and the winner pays the maximum between the bid of the second highest bidder in the auction and the reserve price. In order to determine the winner for CPC and CPA clients, where the pay-off to the publisher is conditioned on a user action, the bids get converted in expected values (also known as eCPMs) using click and conversion rate (CR) prediction models.

The focus of this paper is on improving the performance of a bidder, defined as an agent that takes the CPC or CPA that the  advertiser  is  willing  to  pay  and  submits  a  CPM
bid  for  the  impression. For CPA clients, this bidder takes as input the CPA, that is the value of the sale for the advertiser, computes a predicted CR and produces a bid.
One important aspect of the marketplace is that the numeric range of the possible CPAs is large and depends on the economic value of the sale. The resulting eCPMs vary from ones based on expectations over sales that are frequent and low-value (e.g. song downloads) and ones that are rare and high-value (e.g. hotel reservations). 
An improvement in prediction performance on high CPA sales has a bigger impact on the revenue than a similar improvement that affects low CPA traffic. To take this into account during evaluation, recently proposed metrics on bidding performance make use of the associated CPAs~\cite{hummel2013loss, chapelle2015offline}. 

In this paper, we investigate a novel way of taking into account the sales' CPAs in our CR prediction model for bidding in online advertising auctions, thus bridging the gap between the recently proposed business-aware offline metrics and the current state-of-the-art CR prediction models. The outline is as follows. In Section~\ref{sec:setting}, we present the setting, i.e., our state-of-the-art CR model, the bidder around it, and the recent business-aware offline metrics. Then, in Section~\ref{sec:method}, we introduce our method for taking into account the advertisers' CPAs when training our CR model, which is based on our analysis of the relationship between the Utility loss \cite{hummel2013loss} and the standard log loss. Finally, in Section~\ref{sec:results}, we present our experimental results, both offline and online, before concluding in Section~\ref{sec:conclusion}.

\section{Setting}
\label{sec:setting}

In this section, we discuss the setting of our method. We define the following notations: let $y_i$ be the binary outcome variable indicating if there was a sale or not, $\x_i$ the input display features vector, $c_i$ the display cost, $v_i$ the value of a conversion ---the CPA the advertiser provided--- and $N$ the size of the dataset.

\subsection{Logistic regression for CR modelling}

Current state-of-the-art CR prediction methods range from logistic regression \cite{chapelle2014simple, mcmahan2013ad}, to log-linear models \cite{agarwal2010estimating}, to a combination of log-linear models with decision trees \cite{he2014practical}, and to combining pure response rate prediction with ad ranking \cite{li2015click}. 
In this paper, we use the logistic regression approach from \cite{chapelle2014simple} because of the confirmed state-of-the-art results on click prediction together with the relative ease of implementation and the fact that the model learning can be parallelized efficiently. In this case, the objective function to optimize is the $L_2$ regularized logistic loss:\footnote{While writing down the logistic loss, we assume that the labels are in $\{-1,1\}$ instead of $\{0,1\}$.}
\begin{eqnarray}
\arg\min_\w \sum_{i=1}^N \log(1+\exp(-y_i\w\cdot\x_i)) + \frac{\lambda}{2}\|\w\|^2,
\end{eqnarray} 
with $\lambda$ a regularization value to be tuned.

\subsection{Bidder}

We review in this section the way in which the probabilities of clicks and conversions are used for bidding in an online advertiser auction. The setting is as follows. A {\em bidder} is an agent that competes for an impression that needs to submit a CPM bid to a RTB platform for that impression. It values a certain action---click or conversion---at a certain value $v$ and estimates the probability of the user performing that action if the ad is displayed to be $p$. The value of the impression is thus $p\times v$ and since most RTB platforms rely on second prices auctions, the bidder uses that value for its bid.

Let $c$ be the highest competing bid in that auction. If that value is smaller than the bid $p\times v$, the bidder wins the auction and pays the second price $c$. The payoff of the auction can thus be written as:
\begin{equation}
\left\{\begin{array}{rl}
y\times v - c & \text{if}~p\times v > c \\
0 & \text{otherwise}
\end{array}\right.
\label{eq:empiricalutility}
\end{equation}

\subsection{Offline Metrics}
Let us now detail the recent business-aware offline metrics (Weighted MSE, Utility) that approximate the business impact of a model change.

\paragraph{Weighted MSE}

For a CR model, the classical mean squared error (MSE) can be interpreted as the offline metric that penalizes the volume of poorly explained observed sales. This metric can be extended to weight the display-level squared error with the CPA of the corresponding advertiser and to therefore penalize the model proportionally to the unexplained revenue\footnote{Because CPA times sales is commensurate to revenue.}--- thus yielding a \emph{Weighted MSE} (MSEW). 


\begin{equation}
\textbf{MSEW} = \frac{1}{N}\sum_{i}((y_i - p(\x_i)) \cdot v_i)^2 \\
\label{eq:msew}
\end{equation}


\paragraph{Utility} 
On the other hand, the \emph{Utility}\footnote{This metric is called {\em expected} Utility in \cite{chapelle2015offline}, but we refer to it as Utility in this paper.} metric \cite{chapelle2015offline} allows to model offline the potential change in profit due to a prediction model change. Since the observed profit in historical data is fixed, this metric makes the assumption that the display costs are determined by the highest second bids coming from a second price auction and that they are generated according to a distribution conditioned on the observed display cost:
\begin{equation}
\textbf{Utility} = \sum_{i}\int_{0}^{p(\x_i)v_i} (y_i \cdot v_i - \tilde{c})\Pr(\tilde{c} \mid c_i) d\tilde{c} \\
\label{eq:eu}
\end{equation}

The distribution $\Pr(\tilde{c} \mid c)$ specifies what could have been the second price instead of the observed cost $c$; \cite{chapelle2015offline} suggests a Gamma distribution with $\alpha=\beta c + 1$  and free parameter $\beta$. The motivation for selecting this distribution is that it interpolates nicely between two limit distributions: a Dirac distribution centered at $c$ (as $\beta \to +\infty$) and an improper uniform distribution (as $\beta \to 0$). The former limit case boils down to the {\em empirical} utility \eqref{eq:empiricalutility} while the latter is equivalent to the weighted MSE \eqref{eq:msew} \cite[Theorem 2]{hummel2013loss}. 
The reason for using a distribution around the observed price cost $c$ is that it allows us to penalize model overpredictions on historical data (since all predictions that go over the second price receive the same reward under a utility metric formulation with deterministic cost).


\section{Method}
\label{sec:method}

As we have seen, there is a discrepancy between the CPA-aware offline metrics and the standard loss functions of the current models, such as the log loss function optimized in the logistic regression. This is suboptimal, as current state-of-the-art models for online bidding suffer from misspecification\footnote{A regression model is considered \emph{misspecified} when one of the variables is correlated with the error term, both due to omitted variables bias and due to functional form misspecification.} \cite{hummel2013loss}. We propose the following method to solve this problem.

\subsection{Connection between Utility and Weighted Log Loss}
\label{sec:utilitylossrel}

To the best of our knowledge, the only solution to this problem was proposed in \cite{hummel2013loss}, where the authors design a specific loss function (the \emph{Utility loss}) that take into account the bidder economic performance and which inspired the work on the \emph{Utility metric} \cite{chapelle2015offline}. However, the Utility loss is non-convex as shown in Figure 1 of \cite{hummel2013loss}. We start by investigating the relationship between the Utility loss and the standard log loss, which is used for training current state-of-the-art CR models, in order to determine whether we could extend the standard log loss to solve the problem at hand.

There are several choices to model the distribution of the highest competing bid $\Pr(\tilde{c} \mid c)$ 
in \eqref{eq:eu}. A common distribution mentioned in \cite{hummel2013loss} is the log-normal distribution,  as it nicely captures the fact that the uncertainty in the highest competing bid should be {\em relative} to the specific bid. We will use this distribution in this section as it makes the analysis easier.
Let $\sigma^2$ be the fixed variance of the log normal distribution and $\mu = \log(c)-\sigma^2/2$ chosen in such a way that the mean value is $c$:
$$\Pr(\tilde{c} \mid c, \sigma) = \frac{1}{\sqrt{2\pi}\tilde{c}\sigma} \exp\left(-\frac{(\log (\tilde{c}/c) + \sigma^2/2)^2}{2\sigma^2}\right).$$

The {\em Utility loss} is defined as the opposite of the expected Utility:
$$\ell_\sigma(p, y, v, c) :=  \int_0^{pv} (\tilde{c}-yv) \Pr(\tilde{c} \mid c, \sigma) d\tilde{c}.$$
Of course we cannot make a general connection between the Utility loss and the log loss since the highest competing bid $c$ is involved in the definition of the former, but not in the latter. We can however analyze the behavior of the Utility loss when $c$ is close to our bid $pv$. Note that we expect most of the auctions to be in that regime.
First of all, the derivative of the loss with respect to the prediction $p$ is:
$$\frac{\partial \ell}{\partial p} = v^2(p-y) \Pr(\tilde{c}=pv \mid c, \sigma).$$

Assuming the highest competing bid is equal to our bid, (i.e. $c=pv$), and combining the two previous equations, we get:
$$\frac{\partial \ell}{\partial p} = v^2(p-y) \times \frac{1}{pv} \times \frac{\exp(-\sigma^2/8)}{\sqrt{2\pi}\sigma}
\propto \frac{v(p-y)}{p}.$$

%
Let us now compare the derivatives of the Utility loss and the log loss under the additional assumptions that the probabilities are small ($p\ll 1$), which is typically the case in display advertising.

\begin{center}
\begin{tabular}{l|cc}
& $y=0$ & $y=1$ \\ \hline
Log loss & $\frac{1}{1-p} \approx 1$ & $-\frac{1}{p}$ \\
Utility loss & $v$ & $\frac{v(p-1)}{p} \approx -\frac{v}{p}$
\end{tabular}
\end{center}

We observe that the derivatives are approximately equal, up to a factor $v$. This result motivates the idea of weighting the log loss with the value associated with the sale that we are trying to predict in order to better align the loss used during training and the offline metrics. This can be seen as an extension of the earlier result of \cite{hummel2013loss} which shows that under a uniform distribution of the largest opponent bid, the Utility loss is equivalent to the squared loss weighted by the value. Of course, this approximation works only as long as $c \approx pv$. If this is not the case, directly optimizing the Utility could lead to better performances.

\subsection{Weighted Log Loss} 
As a result of our findings from Section~\ref{sec:utilitylossrel}, we introduce a weighted negative log likelihood (denoted as WNLL) in the context of online bidding and study its behavior. We define: 

\begin{eqnarray}
\text{WNLL} = \sum_{i=1}^N v_i \log(1+\exp(-y_i\w\cdot\x_i)) + \frac{\lambda}{2}\|\w\|^2,
\end{eqnarray} 
where $N$ is the size of the dataset, $\x_i$ is the feature vector, $\w$ is the parameters vector, $y_i$ is the true outcome and $\lambda$ is the hyper-parameter that controls the importance of the $L_2$ regularization factor. Each display is weighted by $v_i$, the CPA of the advertiser associated with the $i^{th}$ display. This is equivalent with generating a dataset where the examples from each advertiser are re-sampled proportionally with its CPA, but with the advantage of not incurring an increase in storage and processing time.

\paragraph{Relationship with Utility loss}

\begin{figure}[h!]
	\includegraphics[width=\columnwidth]{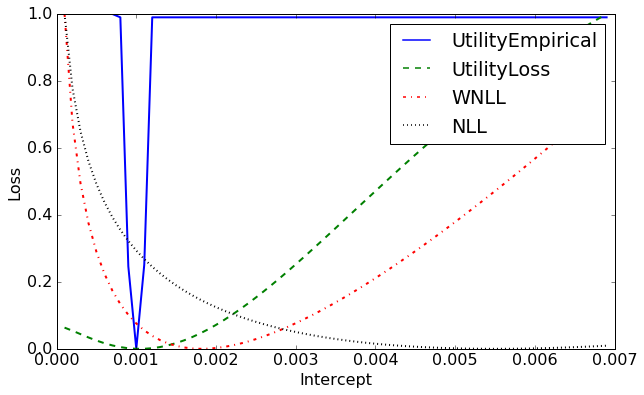}
    \centering
    \caption{Weighted log loss vs. log loss and Utility loss for an intercept-only model on a synthetic dataset with two advertisers.}
    \label{utility_wnll}
\end{figure}

To investigate how this weighting scheme compares to the Utility loss, we use the following toy example.
In Figure~\ref{utility_wnll}, we plot several losses as a function of a fixed predicted conversion rate on an equal mix of two advertisers with very different conversion rates $p_a=0.1\%$ and $p_b=1\%$ (both $p\ll 1$). The CPAs of these advertisers are respectively 50 and 5 and the second prices follow a uniform distribution on $[0.04, 0.06]$. With this setting the advertiser optimal profits and the associated empirical losses are equal and close to zero (to simulate the $c \approx pv$ regime introduced above). 

The Utility loss here and in the rest of this paper is the same as the one defined in \cite{chapelle2015offline} where the highest competing bid $\Pr(\tilde{c} \mid c)$ distribution follows a Gamma distribution with $\alpha = \beta c + 1$ and free parameter $\beta$. 
As $\beta$ goes to infinity, the distribution goes to a Dirac distribution centered around $c$ and the Utility loss converges to the {\em empirical utility} \eqref{eq:empiricalutility}.  We set here $\beta=30$.

The figure shows that the Utility loss has the same optimum point with the empirical utility and that the log loss weighted by CPA (WNLL) has a minimum much closer to the empirical loss $q^*$ that the un-weighted log loss (NLL): $q^*=0.1\%$, $q^*_{WNLL}=0.18\%$, $q^*_{NLL}=0.55\%$.   
 
\begin{table*}[th]
\centering
\begin{tabular}{|l|l|l|l|l|l|l|}
\hline
\multirow{2}{*}{}Weighting & \multicolumn{2}{l|}{$\Delta \text{MSEW}$ (negative is better)}         & \multicolumn{2}{l|}{$\Delta \text{Utility}_{\beta=10}$ (positive is better)} & \multicolumn{2}{l|}{$\Delta \text{Utility}_{\beta=1000}$ (positive is better)} \\ \cline{2-7} 
                           & Train           & Test            & Train          & Test          & Train           & Test           \\ \hline
$CPA$                         & -50.45\% $\pm$ 0.91 & \textbf{-19.57\% $\pm$ 0.65} & 1.44\% $\pm$ 0.02  & \textbf{0.37\% $\pm$ 0.04} & 1.29\% $\pm$ 0.03   & 0.18\% $\pm$ 0.08  \\ \hline
$CPA^\frac{1}{2}$                     & -36.84\% $\pm$ 0.67 & -14.57\% $\pm$ 0.49 & 0.91\% $\pm$ 0.01  & 0.32\% $\pm$ 0.02 & 0.89\% $\pm$ 0.03   & \textbf{0.30\% $\pm$ 0.04}  \\ \hline
$CPA^\frac{1}{4}$                    & -24.54\% $\pm$ 0.46 & -9.26\% $\pm$ 0.3   & 0.51\% $\pm$ 0.01  & 0.18\% $\pm$ 0.01 & 0.51\% $\pm$ 0.02   & 0.19\% $\pm$ 0.03  \\ \hline
\end{tabular}
\caption{Overfitting in WNLL as a function of the CPA weighting scheme. The best performing result for each metric in terms of relative improvement over NLL is indicated in bold. The objective of the two methods is to minimize MSEW (Mean Squared Error weighted by impact on revenue) and to alternatively maximize Utility (a proxy for profit).}
\label{overfitting_uncapped}
\end{table*}

\begin{table*}[ht]
\begin{tabular}{|l|l|}
\hline
$\Lambda$       & $\Delta \text{Utility}_{\beta=1000}$ \\ \hline
$\lambda_h - 40\%$ & 0.34\% $\pm$ 0.09 \\ \hline
$\lambda_h - 20\%$ & 0.32\% $\pm$ 0.05 \\ \hline
$\lambda_h - 10\%$ & 0.29\% $\pm$ 0.05 \\ \hline
$\lambda_h$        & \textbf{0.30\% $\pm$ 0.04} \\ \hline
$\lambda_h + 10\%$ & 0.33\% $\pm$ 0.03 \\ \hline
$\lambda_h + 20\%$ & 0.21\% $\pm$ 0.03 \\ \hline
\end{tabular}
\centering
\caption{Comparison of different values of $\lambda$ around the heuristic value $\lambda_h$ computed by the method covered in Section~\ref{sec:lambda}.}
\label{lambda_study}
\end{table*}

\subsection{Impact of weighting on learning}
In the following, we analyze the impact of moving from a standard log loss to weighted log loss both from the perspective of learning and of regularization.

\paragraph{Learning with importance weights}
\label{sec:importanceweights}
We analyze the learning setup proposed in \cite{chapelle2014simple} where limited memory BFGS \cite{liu1989limited,nocedal1980updating} (L-BFGS) is warm-started with stochastic gradient descent \cite{bottou2010large} (SGD). For both algorithms, we multiply the gradient of the loss of each example by $v$, where $v$ is the weight associated with the example. This is straightforward to implement. Note that exact importance weighting for SGD is more tricky and that dedicated weighting schemes exist \cite{karampatziakis2010online}. However, in our case, we use SGD only for warm-starting L-BFGS and our approximate method of including importance weights proved  to be sufficient.

\paragraph{Impact on the regularization parameter}
\label{sec:lambda}

In the case of switching from log loss to the weighed log loss, the value of the $\lambda$ hyper-parameter for NLL needs to be adapted to WNLL. To do that, we use the following simple rule that adapts $\lambda$ depending on the value of the importance weights used, i.e. of the average CPA of each advertiser:
\begin{eqnarray}
\lambda_{WNLL} = \lambda_{NLL} \times \frac{\sum_i v_i}{N}
\end{eqnarray}

For our experiments we use the following heuristic to set the value of $\lambda$ in the un-weighted case, as suggested in \cite{joachims1999making,chapelle2014modeling}:
\begin{eqnarray}
\lambda_{NLL}^h = \frac{1}{N} \sum_{i=1}^{N} \|\x_i\|_{2}^{2}
\end{eqnarray}
where $x_i$ represents a training instance vector and $n$ is the size of the dataset. 

We show how well the lambda rescaling scheme performs relative to other values of $\lambda$ in the context of $L_2$ regularization in Table~\ref{lambda_study} in Section~\ref{sec:results}.


\begin{table*}[ht]
\begin{tabular}{|l|l|l|l|}
\hline
Metric & Global          & HighCPA(\textgreater10) & HighCPA(\textgreater10), LowSales(\textless30) \\ \hline
$\Delta \text{MSEW}$   & -14.57\% $\pm$ 0.49 & -15.26\% $\pm$ 0.53         & -22.36\% $\pm$ 1.02                                \\ \hline
$\Delta \text{Utility}_{\beta=10}$    & 0.32\% $\pm$ 0.02   & 0.78\% $\pm$ 0.06           & 1.70\% $\pm$ 0.16                                  \\ \hline
$\Delta \text{Utility}_{\beta=1000}$  & 0.30\% $\pm$ 0.04   & 0.78\% $\pm$ 0.11           & 1.72\% $\pm$ 0.21                                  \\ \hline
\end{tabular}
\centering
\caption{Relative improvement of WNLL vs. NLL on the open Criteo dataset, both globally and for advertising campaigns with high CPA.}
\label{open_dataset}
\end{table*}

\section{Experiments}
\label{sec:results}
In this section, we present our experimental results when applying our method to improve a state-of-the-art conversion-rate prediction model for bidding in online advertising auctions. We present offline results followed by online experiments on live traffic.

\subsection{Offline results on public dataset}

For comparing WNLL and NLL, we used a public dataset released by Criteo as supporting material of \cite{chapelle2014modeling}. The dataset contains a sample of post-click conversions with a matching window of 30 days. For simplicity, we choose the setup (denoted by the authors as the "oracle" setup), where we use for training the entire set of positive examples (clicked displays that converted within the next 30 days) and apply the model without leaving a 30 days window for evaluation, as it would be needed in a live system. 

So far in the paper we have assumed a model predicting the conversions at the display level. Since the records in this public dataset are at the click level, we will instead consider in this section the task of predicting a probability of conversion given click. In order for this to be used in a production system the post-click probability would need to be further multiplied by a probability of click given display, as explained in \cite{chapelle2014modeling}.

Since the objective of the dataset was to show empirically the importance of modeling delayed conversions, the associated display costs and conversion revenue are not included. 
To be able to evaluate our method, we introduce a simplified cost and revenue scheme where all clicks have a constant cost of 1, and for each advertising campaign the CPA is inversely proportional with the historical post-click conversion rate, meaning that each advertising campaign $c$ is assumed to be contributing equally to the overall revenue: 
$cost_{c} = 1, CPA_{c} = \frac{1}{AvgCR_{c}}$ and $AvgCR_c$ is the campaign average CR with smoothing as explained below.

For the experimental results, we compare the results of the baseline log loss (NLL) and the results of the weighted log loss (WNLL). We take a 2 weeks period (weeks 3 and 4) as the test period (each model is trained on a period of up to three weeks and used to predict the conversion rate on next day traffic). For the historical CR estimate, we use the 2 weeks period before the first test day (weeks 1 and 2) to compute average conversion rates for the campaigns. To handle the case of new campaigns appearing during the testing period, we set the final campaign CR estimator to be: $SmoothCR_c = \frac{\#sales_c + AvgCR}{\#clicks_c + 1}$ where $AvgCR$ is the overall average CR computed in the two weeks and is equal to $0.23$. 

\paragraph{Offline metrics}
The evaluation metrics are the MSEW \eqref{eq:msew} and the Utility \eqref{eq:eu} computed using a Gamma distribution with free parameter $\beta$ as in \cite{chapelle2015offline}.
A more accurate model is expected to decrease the MSEW and increase the Utility.
All the results presented in this section are provided with confidence intervals computed using bootstrap.

\paragraph{The impact of CPA weight factor on performance.}
As previously covered in the cost-sensitive learning literature \cite{cortes2010learning}, associating big costs to the training examples can lead to overfitting. We have established experimentally on our internal traffic that the best solution to combat overfitting is reducing the amplitude of the final weights used during model learning by a combination of capping the maximum values of the CPAs and a magnitude reduction function (sqrt) on the CPA. In Table~\ref{overfitting_uncapped} we show the relative performance of various CPA dampening schemes on the open dataset, for a maximum value of CPA =20. We see that the amount of overfitting in terms of MSEW is comparable across all of the three weighting schemes, but that $CPA^\frac{1}{2}$ overfits less in terms of $\text{Utility}_{\beta=1000}$ and  has better performance on the test set.
Another practical benefit of capping the CPA weights is that the CPA estimate can be noisy to start with, especially in the case of new campaigns or of campaigns with low number of sales. Without capping, the model might learn to fit only campaigns with very high CPA and predict with the intercept for the others. Though capping introduces bias in the CPA estimate, as long as the resulting weighting scheme is closer to the actual revenue breakdown over campaigns than the baseline uniform weighting (CPA=1), WNLL will likely outperform NLL.


\paragraph{Impact of $\lambda$}
We benchmark $\lambda$ values around the value $\lambda_h$ produced by the heuristic proposed in Section~\ref{sec:lambda}. We observe in table \ref{lambda_study} that the proposed $\lambda$ is very close to the optimal value. This finding mirrors the results obtained on internal data. For this reason, we use the $\lambda$ proposed by the heuristic in all our experiments.


\paragraph{WNLL vs. NLL on high CPA}
 

In Table~\ref{open_dataset}, we report performance of our best WNLL setup (with weighting scheme $CPA^\frac{1}{2}$) on the evaluation metrics that are closest to the actual business metrics, e.g. MSEW and Utility. We observe that the biggest lift of WNLL vs. NLL is on the advertising campaigns with high CPA and a low number of sales ($<$30) in the period of reference (weeks 1 and 2). This confirms our hypothesis that WNLL should outperform NLL on campaigns with low volume of sales, but high CPA. Because of that, the actual economic impact of switching to WNLL could be even greater, depending of the relative proportion of traffic with high CPA and low number of positives. As we will show next, our online experiments confirm this finding.

\subsection{Online experiments}
We ran an A/B test of the change of the loss function to WNLL in the conversion-rate model. The A/B test was done on more than 1 Billion ad displays, on world-wide traffic. Our change resulted in a +2\% lift in ROI, which is a considerable lift compared to typical improvements in this field. We observed significant savings in display cost, coupled with an increase in sales performance for the advertisers, especially on the campaigns with high CPA and low number of sales that account for a significant proportion of revenue.
In terms of development and operational costs, the change in the loss function took only a couple of weeks to put in production, since the code change is minimal, as shown in Section~\ref{sec:method}. Furthermore, the training time of the model did not change.


\section{Conclusion}
\label{sec:conclusion}
We investigated the relationship between the Utility loss and the standard log loss. This analysis motivated the idea of weighting the log loss with the value associated with the sale that we are trying to predict (CPA) in order to better align the loss used during training and the offline metrics. Then, we presented and analyzed a cost weighting scheme that takes into account the advertisers' CPAs when training a CR model for bidding in online advertising auctions and discussed its impact on learning and regularization. We showed that this cost weighting scheme leads to a loss function whose optimal point is much closer to the optimum point (reached by optimizing the Utility loss) than the one of the standard log loss. We finally demonstrated that our method allows us to improve a state-of-the-art CR prediction model used  for bidding in online advertising auctions and leads to large significant lifts in offline performance (on a public data set) and online performance as evaluated through an A/B test.

\paragraph{Future work} As future work, we plan on investigating two directions. First, optimize directly the Utility loss \cite{hummel2013loss}, which is non-convex and thus requires careful optimization. Second, use a different convex approximation of the Utility loss, as proposed in  \cite[Section 6]{hummel2013loss}.

\section*{Acknowledgments}
We would like to thank Vianney Perchet, Nicolas Le Roux, Olivier Koch, Etienne Sanson, Cyrille Dubarry, Alexandre Gilotte and Dmitry Pavlov for their useful comments on early versions of this paper.

\bibliographystyle{ACM-Reference-Format}
\bibliography{refs}


\begin{thebibliography}{00}


\ifx \showCODEN    \undefined \def \showCODEN     #1{\unskip}     \fi
\ifx \showDOI      \undefined \def \showDOI       #1{{\tt DOI:}\penalty0{#1}\ }
  \fi
\ifx \showISBNx    \undefined \def \showISBNx     #1{\unskip}     \fi
\ifx \showISBNxiii \undefined \def \showISBNxiii  #1{\unskip}     \fi
\ifx \showISSN     \undefined \def \showISSN      #1{\unskip}     \fi
\ifx \showLCCN     \undefined \def \showLCCN      #1{\unskip}     \fi
\ifx \shownote     \undefined \def \shownote      #1{#1}          \fi
\ifx \showarticletitle \undefined \def \showarticletitle #1{#1}   \fi
\ifx \showURL      \undefined \def \showURL       {\relax}        \fi
\providecommand\bibfield[2]{#2}
\providecommand\bibinfo[2]{#2}
\providecommand\natexlab[1]{#1}
\providecommand\showeprint[2][]{arXiv:#2}

\bibitem[\protect\citeauthoryear{??}{sta}{}]%
        {statista}
\bibinfo{title}{Digital advertising spend in 2015}.
\newblock
  \bibinfo{howpublished}{\url{http://www.statista.com/statistics/237974/online-advertising-spending-worldwide/}}.
    (\bibinfo{year}{????}).
\newblock
\newblock
\shownote{Accessed: 2015-10-15.}


\bibitem[\protect\citeauthoryear{Agarwal, Agrawal, Khanna, and Kota}{Agarwal
  et~al\mbox{.}}{2010}]%
        {agarwal2010estimating}
\bibfield{author}{\bibinfo{person}{Deepak Agarwal}, \bibinfo{person}{Rahul
  Agrawal}, \bibinfo{person}{Rajiv Khanna}, {and} \bibinfo{person}{Nagaraj
  Kota}.} \bibinfo{year}{2010}\natexlab{}.
\newblock \showarticletitle{Estimating rates of rare events with multiple
  hierarchies through scalable log-linear models}. In \bibinfo{booktitle}{{\em
  Proceedings of the 16th ACM SIGKDD international conference on Knowledge
  discovery and data mining}}. ACM, \bibinfo{pages}{213--222}.
\newblock


\bibitem[\protect\citeauthoryear{Bottou}{Bottou}{2010}]%
        {bottou2010large}
\bibfield{author}{\bibinfo{person}{L{\'e}on Bottou}.}
  \bibinfo{year}{2010}\natexlab{}.
\newblock \showarticletitle{Large-scale machine learning with stochastic
  gradient descent}.
\newblock In \bibinfo{booktitle}{{\em COMPSTAT '10}}.
  \bibinfo{publisher}{Springer}, \bibinfo{pages}{177--186}.
\newblock


\bibitem[\protect\citeauthoryear{Chapelle}{Chapelle}{2014}]%
        {chapelle2014modeling}
\bibfield{author}{\bibinfo{person}{Olivier Chapelle}.}
  \bibinfo{year}{2014}\natexlab{}.
\newblock \showarticletitle{Modeling delayed feedback in display advertising}.
  In \bibinfo{booktitle}{{\em Proceedings of the 20th ACM SIGKDD international
  conference on Knowledge discovery and data mining}}. ACM,
  \bibinfo{pages}{1097--1105}.
\newblock


\bibitem[\protect\citeauthoryear{Chapelle}{Chapelle}{2015}]%
        {chapelle2015offline}
\bibfield{author}{\bibinfo{person}{Olivier Chapelle}.}
  \bibinfo{year}{2015}\natexlab{}.
\newblock \showarticletitle{Offline Evaluation of Response Prediction in Online
  Advertising Auctions}. In \bibinfo{booktitle}{{\em Proceedings of the 24th
  International Conference on World Wide Web Companion}}. International World
  Wide Web Conferences Steering Committee, \bibinfo{pages}{919--922}.
\newblock


\bibitem[\protect\citeauthoryear{Chapelle, Manavoglu, and Rosales}{Chapelle
  et~al\mbox{.}}{2014}]%
        {chapelle2014simple}
\bibfield{author}{\bibinfo{person}{Olivier Chapelle}, \bibinfo{person}{Eren
  Manavoglu}, {and} \bibinfo{person}{Romer Rosales}.}
  \bibinfo{year}{2014}\natexlab{}.
\newblock \showarticletitle{Simple and scalable response prediction for display
  advertising}.
\newblock \bibinfo{journal}{{\em ACM Transactions on Intelligent Systems and
  Technology (TIST)\/}} \bibinfo{volume}{5}, \bibinfo{number}{4}
  (\bibinfo{year}{2014}), \bibinfo{pages}{61}.
\newblock


\bibitem[\protect\citeauthoryear{Cortes, Mansour, and Mohri}{Cortes
  et~al\mbox{.}}{2010}]%
        {cortes2010learning}
\bibfield{author}{\bibinfo{person}{Corinna Cortes}, \bibinfo{person}{Yishay
  Mansour}, {and} \bibinfo{person}{Mehryar Mohri}.}
  \bibinfo{year}{2010}\natexlab{}.
\newblock \showarticletitle{Learning bounds for importance weighting}. In
  \bibinfo{booktitle}{{\em Advances in neural information processing systems}}.
  \bibinfo{pages}{442--450}.
\newblock


\bibitem[\protect\citeauthoryear{He, Pan, Jin, Xu, Liu, Xu, Shi, Atallah,
  Herbrich, Bowers, et~al\mbox{.}}{He et~al\mbox{.}}{2014}]%
        {he2014practical}
\bibfield{author}{\bibinfo{person}{Xinran He}, \bibinfo{person}{Junfeng Pan},
  \bibinfo{person}{Ou Jin}, \bibinfo{person}{Tianbing Xu}, \bibinfo{person}{Bo
  Liu}, \bibinfo{person}{Tao Xu}, \bibinfo{person}{Yanxin Shi},
  \bibinfo{person}{Antoine Atallah}, \bibinfo{person}{Ralf Herbrich},
  \bibinfo{person}{Stuart Bowers}, {and} \bibinfo{person}{others}.}
  \bibinfo{year}{2014}\natexlab{}.
\newblock \showarticletitle{Practical lessons from predicting clicks on ads at
  facebook}. In \bibinfo{booktitle}{{\em Proceedings of 20th ACM SIGKDD
  Conference on Knowledge Discovery and Data Mining}}. ACM,
  \bibinfo{pages}{1--9}.
\newblock


\bibitem[\protect\citeauthoryear{Hummel and McAfee}{Hummel and McAfee}{2013}]%
        {hummel2013loss}
\bibfield{author}{\bibinfo{person}{Patrick Hummel} {and}
  \bibinfo{person}{R~Preston McAfee}.} \bibinfo{year}{2013}\natexlab{}.
\newblock \showarticletitle{Loss functions for predicted click through rates in
  auctions for online advertising}.
\newblock \bibinfo{journal}{{\em Preprint, Google Inc\/}}
  (\bibinfo{year}{2013}).
\newblock


\bibitem[\protect\citeauthoryear{Joachims}{Joachims}{1999}]%
        {joachims1999making}
\bibfield{author}{\bibinfo{person}{Thorsten Joachims}.}
  \bibinfo{year}{1999}\natexlab{}.
\newblock \bibinfo{booktitle}{{\em Making large scale SVM learning practical}}.
\newblock \bibinfo{type}{{T}echnical {R}eport}.
  \bibinfo{institution}{Universit{\"a}t Dortmund}.
\newblock


\bibitem[\protect\citeauthoryear{Karampatziakis and Langford}{Karampatziakis
  and Langford}{2010}]%
        {karampatziakis2010online}
\bibfield{author}{\bibinfo{person}{Nikos Karampatziakis} {and}
  \bibinfo{person}{John Langford}.} \bibinfo{year}{2010}\natexlab{}.
\newblock \showarticletitle{Online importance weight aware updates}.
\newblock \bibinfo{journal}{{\em arXiv preprint arXiv:1011.1576\/}}
  (\bibinfo{year}{2010}).
\newblock


\bibitem[\protect\citeauthoryear{Li, Lu, Mei, Wang, and Pandey}{Li
  et~al\mbox{.}}{2015}]%
        {li2015click}
\bibfield{author}{\bibinfo{person}{Cheng Li}, \bibinfo{person}{Yue Lu},
  \bibinfo{person}{Qiaozhu Mei}, \bibinfo{person}{Dong Wang}, {and}
  \bibinfo{person}{Sandeep Pandey}.} \bibinfo{year}{2015}\natexlab{}.
\newblock \showarticletitle{Click-through Prediction for Advertising in Twitter
  Timeline}. In \bibinfo{booktitle}{{\em Proceedings of the 21th ACM SIGKDD
  International Conference on Knowledge Discovery and Data Mining}}. ACM,
  \bibinfo{pages}{1959--1968}.
\newblock


\bibitem[\protect\citeauthoryear{Liu and Nocedal}{Liu and Nocedal}{1989}]%
        {liu1989limited}
\bibfield{author}{\bibinfo{person}{Dong~C Liu} {and} \bibinfo{person}{Jorge
  Nocedal}.} \bibinfo{year}{1989}\natexlab{}.
\newblock \showarticletitle{On the limited memory BFGS method for large scale
  optimization}.
\newblock \bibinfo{journal}{{\em Mathematical programming\/}}
  \bibinfo{volume}{45}, \bibinfo{number}{1-3} (\bibinfo{year}{1989}),
  \bibinfo{pages}{503--528}.
\newblock


\bibitem[\protect\citeauthoryear{McMahan, Holt, Sculley, Young, Ebner, Grady,
  Nie, Phillips, Davydov, Golovin, et~al\mbox{.}}{McMahan
  et~al\mbox{.}}{2013}]%
        {mcmahan2013ad}
\bibfield{author}{\bibinfo{person}{H~Brendan McMahan}, \bibinfo{person}{Gary
  Holt}, \bibinfo{person}{David Sculley}, \bibinfo{person}{Michael Young},
  \bibinfo{person}{Dietmar Ebner}, \bibinfo{person}{Julian Grady},
  \bibinfo{person}{Lan Nie}, \bibinfo{person}{Todd Phillips},
  \bibinfo{person}{Eugene Davydov}, \bibinfo{person}{Daniel Golovin}, {and}
  \bibinfo{person}{others}.} \bibinfo{year}{2013}\natexlab{}.
\newblock \showarticletitle{Ad click prediction: a view from the trenches}. In
  \bibinfo{booktitle}{{\em Proceedings of the 19th ACM SIGKDD international
  conference on Knowledge discovery and data mining}}. ACM,
  \bibinfo{pages}{1222--1230}.
\newblock


\bibitem[\protect\citeauthoryear{Nocedal}{Nocedal}{1980}]%
        {nocedal1980updating}
\bibfield{author}{\bibinfo{person}{Jorge Nocedal}.}
  \bibinfo{year}{1980}\natexlab{}.
\newblock \showarticletitle{Updating quasi-Newton matrices with limited
  storage}.
\newblock \bibinfo{journal}{{\em Mathematics of computation\/}}
  \bibinfo{volume}{35}, \bibinfo{number}{151} (\bibinfo{year}{1980}),
  \bibinfo{pages}{773--782}.
\newblock


\bibitem[\protect\citeauthoryear{Yuan, Wang, and Zhao}{Yuan
  et~al\mbox{.}}{2013}]%
        {yuan2013real}
\bibfield{author}{\bibinfo{person}{Shuai Yuan}, \bibinfo{person}{Jun Wang},
  {and} \bibinfo{person}{Xiaoxue Zhao}.} \bibinfo{year}{2013}\natexlab{}.
\newblock \showarticletitle{Real-time bidding for online advertising:
  measurement and analysis}. In \bibinfo{booktitle}{{\em Proceedings of the
  Seventh International Workshop on Data Mining for Online Advertising}}. ACM,
  \bibinfo{pages}{3}.
\newblock


\end{thebibliography}

\end{document}